MVMGNN: Multi-View Masked Graph Neural Network for Alzheimer's Disease Diagnosis using Structural MRI

Ni Yao[1] ·Zhenxu Wang[1]·Danyang Sun[1]·Chuang Han[1]·Yanting Li[1]·Jiaofen Nan[1]·Fubao Zhu[1,*] ·Chen Zhao[2*]·Weihua Zhou[3,4]

[1] School of Computer Science and Technology, Zhengzhou University of Light Industry, Zhengzhou 450002, Henan, China

[2] Department of Computer Science, Kennesaw State University Marietta, GA, USA

[3] Department of Applied Computing, Michigan Technological University, Houghton, MI, USA

[4] Center for Biocomputing and Digital Health, Institute of Computing and Cybersystems, and Health Research Institute, Michigan Technological University, Houghton, MI, USA

Zhenxu Wang and Chen Zhao contributed equally.

[*] Correspondence:

Fubao Zhu

Email address: fbzhu@zzuli.edu.cn

Mailing address: School of Computer Science and Technology, Zhengzhou University of Light Industry, Zhengzhou 450002, Henan, China

Chen Zhao

Email address: czhao4@kennesaw.edu

Mailing address: 680 Arntson Dr, Marietta, GA 30060

**Abstract**

Alzheimer's disease (AD) is a common neurodegenerative disorder, and early diagnosis is of great significance for delaying disease progression and enabling timely intervention. Mild cognitive impairment (MCI), which represents an intermediate clinical stage between cognitively normal aging and AD, is also an important target for early identification because individuals with MCI are at an increased risk of progressing to AD. Structural magnetic resonance imaging (sMRI) provides detailed characterization of anatomical structures and plays an important role in AD-related brain analysis. However, existing sMRI-based brain network methods typically rely on a single graph construction strategy, limiting their ability to jointly capture spatial relationships and morphological similarities between brain regions. While multi-view approaches can incorporate complementary information, they are often affected by redundant features and noisy connections without effective selection and fusion mechanisms. To address these issues, this paper proposes an sMRI-based multi-view masked graph neural network model (MVMGNN) for AD diagnosis. Brain regions are regarded as nodes with and radiomics features, and two complementary graph views are constructed based on spatial proximity and feature similarity. A joint node–edge masking mechanism is proposed to simultaneously select radiomics feature dimensions and structural connections, reducing redundancy during graph learning. Furthermore, a patient-level cross-view gated fusion mechanism is proposed to integrate multi-view representations. Experimental results on the ADNI dataset demonstrate that MVMGNN outperforms several competing approaches in AD classification. Interpretability analysis further demonstrates that MVMGNN is able to identify key brain regions associated with AD, providing useful insights into discriminative patterns in sMRI-based brain networks.Our implementation is publicly available at https://github.com/chenzhao2023/MVMGNN_AD

## 1. Introduction

Alzheimer’s disease (AD) is a chronic neurodegenerative disorder commonly observed in the elderly population and one of the leading causes of dementia. With the acceleration of global population aging, AD has become one of the most prevalent, fatal, and socioeconomically burdensome diseases of the 21st century[1]. Therefore, achieving early and accurate diagnosis of AD is of great practical significance for delaying disease progression and developing individualized intervention strategies[2].

In recent years, computer-aided diagnosis methods based on neuroimaging data have made significant progress in the prediction of AD and mild cognitive impairment (MCI)[3]. Among various neuroimaging modalities, structural magnetic resonance imaging (sMRI) enables detailed characterization of brain anatomical structures and regional morphological features, making it an important imaging modality for investigating AD-related structural abnormalities[4][5].

Existing sMRI-based AD studies have commonly represented brain images as feature vectors. Regional morphological measurements, such as brain volume, cortical thickness, and gray matter density, have been widely used to characterize AD-related structural atrophy[4]. Voxel-based morphometry has also been employed to analyze whole-brain gray matter differences and support MRI-based AD classification[6][7]. With the development of radiomics, high-dimensional quantitative features have been extracted from sMRI to describe intensity distribution, shape properties, and texture patterns within brain regions[8]. These MRI texture and radiomics features have shown potential for MCI conversion prediction and early AD identification[9][10]. However, these feature-vector-based methods mainly focus on local structural alterations and insufficiently model organizational relationships among brain regions.

Brain network modeling provides a natural graph-based representation for inter-regional relationships. In this representation, brain regions are treated as nodes, and associations between brain regions are modeled as edges. For example, large-scale cortical network analysis has been used to reveal abnormal topological patterns in AD[11], single-subject gray matter graphs have been constructed to characterize individual-level structural organization[12], and morphological brain networks based on multiple structural features have been applied to AD and MCI identification[13]. Graph neural networks (GNNs) are suitable for such graph-structured data because they can jointly learn node attributes and topological structures through message passing and neighborhood aggregation[14]. Several GNN-based studies have further demonstrated their potential in neurological disorder diagnosis and AD-related graph learning tasks[15][16][17][18].

However, existing sMRI-based graph learning methods still have certain limitations. First, several representative studies construct brain graphs based on a single predefined relationship. For example, Tijms et al. constructed single-subject gray matter graphs to quantify cortical structural organization in AD[12], and Yu et al. proposed individual morphological brain networks based on multivariate Euclidean distances between brain regions[19]. These methods demonstrate the value of graph-based structural representation, but they mainly rely on a single graph construction criterion and do not fully exploit heterogeneous information from different graph views. To address this issue, we introduce two views that contain different types of connectivity information. Specifically, the spatial proximity view describes local anatomical neighborhood relationships among brain regions, whereas the feature similarity view represents morphological relationships in the feature space. By constructing brain graphs from different perspectives, these two views provide complementary information for structural brain network modeling.

On the other hand, multi-feature, multilayer, or multi-view brain networks have also been explored. Zheng et al. constructed brain networks using multiple morphological features for AD and MCI identification[13], Mahjoub et al. proposed morphological brain multiplexes to characterize dementia-related connectional biomarkers[20], and Wang et al. introduced a multi-view graph representation learning method to capture information from different brain network perspectives[21]. However, multi-view brain network modeling may still involve heterogeneous views, noisy edges, and data inconsistency. If different views are directly integrated without effective feature selection, redundant features and weakly relevant connections may interfere with graph representation learning. To address this issue, we propose a multi-view masking and gated fusion strategy that adaptively selects discriminative node features and structural connections within each view, while the global gating vector is used to integrate multi-view representations at the cross-view level. In this way, the model can reduce the influence of redundant information across heterogeneous views.

To address these issues, we propose a Multi-View Masked Graph Neural Network (MVMGNN) for AD stage classification. The main contributions are summarized as follows:

(1) We propose a dual-view brain network modeling strategy that captures complementary inter-regional relationships by jointly considering spatial proximity and feature similarity. This enables a more comprehensive representation of structural brain organization compared to conventional single-view graph construction methods.

(2)We introduce a unified masking framework that simultaneously performs feature selection at the node level and connection selection at the edge level within each graph view. This mechanism effectively suppresses redundant radiomics features and noisy or weakly relevant connections during graph learning.

(3)We develop a patient-level gated fusion module to adaptively integrate heterogeneous information from multiple graph views, allowing the MVMGNN to dynamically weight different views and improves robustness against view inconsistency.

(4)Experimental results on AD-related classification tasks demonstrate that MVMGNN achieves superior overall performance over multiple comparison methods across several evaluation metrics.

## 2. Method

Figure 1 illustrates the overview of the proposed MVMGNN. The model takes sMRI images as input. First, ROI-level node attributes are obtained through brain region segmentation and feature extraction, and multiple structural brain network views are generated based on different graph construction strategies. Subsequently, a joint node–edge masking mechanism is introduced within each view to explicitly modulate node features and edge connections, thereby enabling intra-view structural selection. The masked views are then separately fed into GCN encoders for representation learning, followed by graph-level pooling to obtain view-level representations. On this basis, the cross-view gated fusion module adaptively integrates representations from different views at the sample level, and the final classification result is generated through a fully connected classification layer.

**Figure 1.** Overview of the proposed MVMGNN.

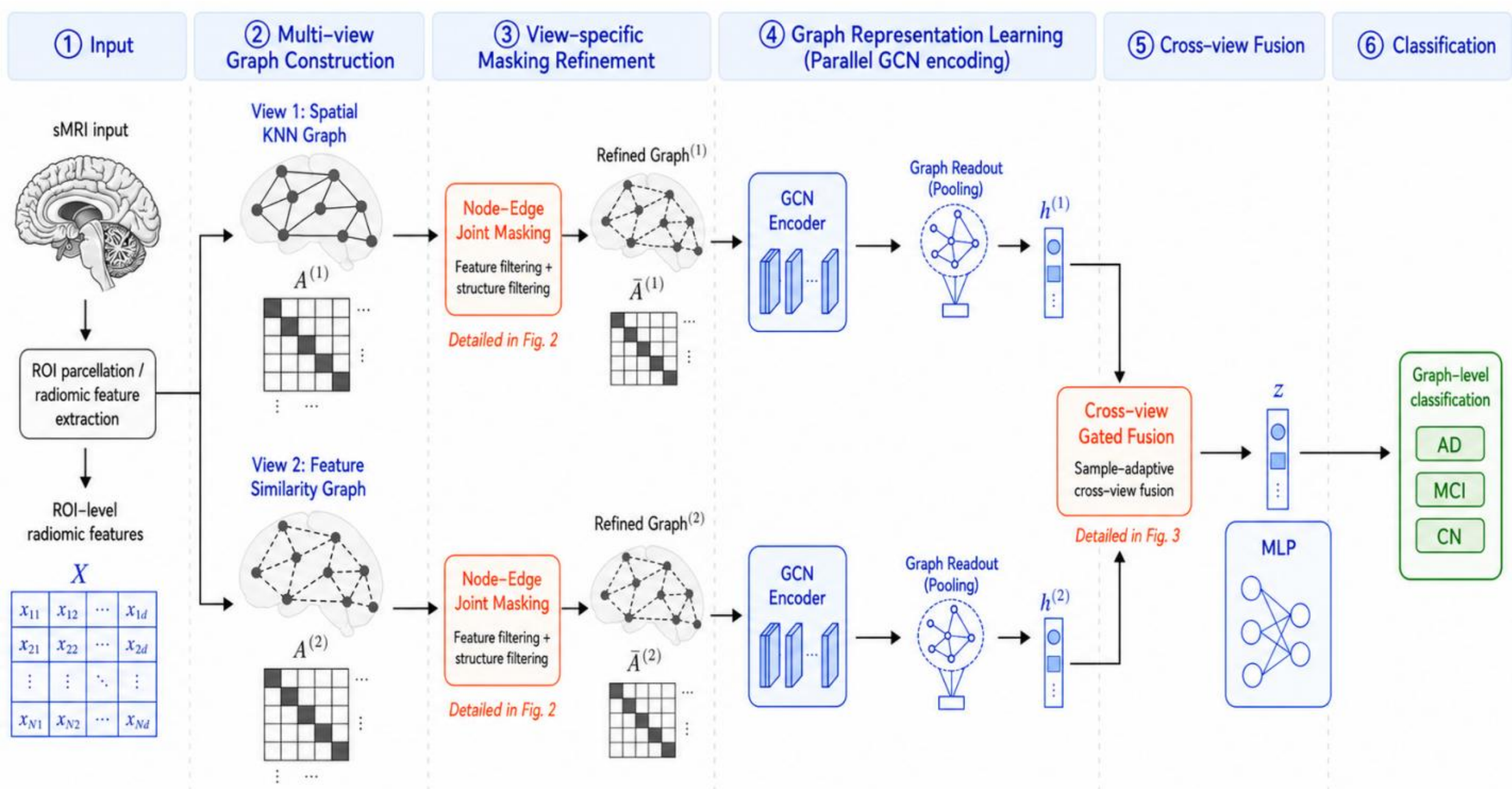

## 2.1 Brain Graph Construction

To obtain reliable region-level structural features, each T1-weighted sMRI scan in this study was automatically processed using FreeSurfer[22]. In detail, the segmentation result combines cortical parcellation based on the Desikan–Killiany (DK) atlas with subcortical structure segmentation, containing a total of 104 non-zero labels corresponding to 104 brain regions[23]. On this basis, radiomics features were extracted from each brain region using PyRadiomics[24], including first-order intensity statistics, shape-based features, and texture features derived from gray-level matrices. Finally, a radiomics feature vector was obtained for each brain region, and invalid feature values were uniformly set to 0. As a result, each sMRI scan was represented by a feature matrix $X \in R^{N \times F}$:

$$X = \begin{bmatrix} feat_{1,1} & \cdots & feat_{1,F} \\ \vdots & \ddots & \vdots \\ feat_{N,1} & \cdots & feat_{N,F} \end{bmatrix} \quad (1)$$

where N denotes the number of brain regions and F denotes the dimensionality of the radiomics feature vector. The matrix element $feat_{n,k}$ denotes the n-th brain region of the k-th radiomics feature.

In this study, N=104 and F=113, indicating that each brain image was parcellated into 104 brain regions, with a 113-dimensional radiomics feature vector extracted from each region.

When constructing the structural brain network, this study adopts two complementary edge construction strategies to characterize potential associations among brain regions from two different perspectives: spatial structural relationships and feature similarity. In detail, each sMRI sample is represented as a multi-view undirected graph $G = (V, X, A^K, A^S)$, where $V=v_1,v_2,\cdots,v_N$ denotes the set of brain region nodes, X denotes the node feature matrix, $A^K$ is the spatial connectivity and $A^S$ is the feature similarity connectivity.

KNN-Based Spatial Connectivity. To characterize the local proximity relationships among brain regions in anatomical space, this study constructs a K-nearest neighbor (KNN) structural graph based on the spatial locations of brain region centroids. The K nearest neighbors determined by the Euclidean distance are then selected, and undirected edges are established between $v_i$ and these neighbors. The resulting spatial connectivity matrix based on proximity distance in the spatial domain, denoted as $A^K \in R^{N\times N}$ :

$$A^K_{(i,j)} = \begin{cases} 1, \text{ if } i\neq j \text{ and } v_i \in N_K(v_j) \\ 0, \text{ otherwise} \end{cases} \quad (2)$$

where $N_K(v_j)$ denotes the set of K nearest neighboring nodes of node vi.

Feature Similarity-Based Graph Construction. In addition to spatial proximity relationships, similarities among brain regions in the structural feature space also contain important discriminative information. For any two brain region nodes $\mathbf{v_i}$ and $\mathbf{v_j}$,their radiomics feature vectors are denoted as $feat_{v_i}=X_{i,:}$ and $feat_{v_j}=X_{j,:}$ . The cosine similarity between their radiomics feature vectors $feat_{v_i}$ and $feat_{v_j}$ is calculated.

$$S_{ij} = \frac{\left(feat_{v_i}{}^T feat_{v_j}\right)}{\left(\|feat_{v_i}\|^2 * \|feat_{v_j}\|^2\right)} \quad (3)$$

where $feat_{v_i}$ and $feat_{v_j}$ denote the complete radiomics feature vectors of brain regions $v_i$ and $v_j$ .

A predefined threshold δ is set, and an edge is established when the feature similarity between two brain regions is greater than this threshold. The final feature similarity-based connectivity matrix $A^S \in R^{N\times N}$ is defined in Eq. 4.

$$A^S_{(i,j)} = \begin{cases} S_{ij}, \text{ if } S_{ij}\geq\delta \\ 0, \text{ otherwise} \end{cases} \quad (4)$$

**2.2 Graph Convolutional Network Module**

A graph convolutional network (GCN) is employed as the basic encoder to jointly model node features and topological structures under each view. The overall architecture consists of three components: graph convolutional layers, a graph-level pooling layer, and a multilayer perceptron (MLP) classifier. The graph convolutional layers are used to learn node representations, the graph-level pooling layer aggregates node representations into a graph-level representation, and the MLP classifier outputs the final classification result based on the graph-level representation.

**2.3 Multi-View Masking Module**

This study introduces a joint node–edge masking mechanism within each view to explicitly modulate node attributes and edge structures before information propagation. The detailed operation of this module is illustrated in Figure 2.

**Figure 2.** Overview of the multi-view masking module for graph feature representation learning

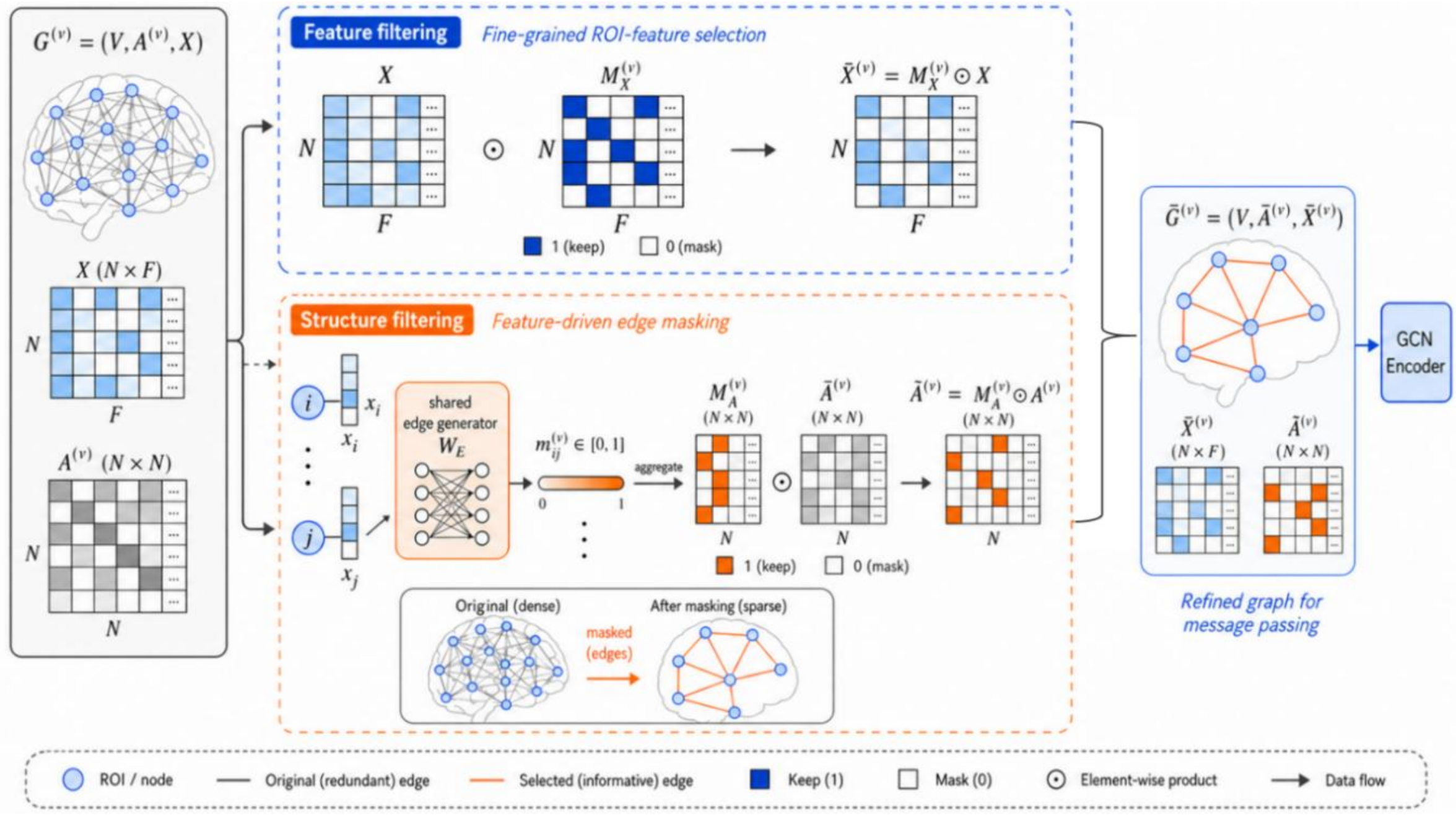

The learnable masking components consist of view-specific feature mask matrices and view-specific edge mask matrices, which are used to select discriminative feature dimensions and key structural connections within each view. Specifically, the node feature matrix X is composed of a N×F radiomics feature matrix, while $A^K$ and $A^S$ are N×N adjacency matrices. The dimensions of the mask matrices are consistent with the corresponding components in the graph $G = (V, X, A^K, A^S)$. During training, the mask matrices are jointly optimized using all training samples. In contrast, the edge masks are dynamically generated from the features of the two endpoint nodes of each candidate edge, enabling the edge connection weights to be modulated according to ROI-specific features.

First, for the node feature matrix X in the γ-th view, where $\gamma \in \{K, S\}$ denotes the view index, a node mask $M_X^{r} \in R^{N\times F}$ with the same dimensionality is learned. This mask is directly applied to individual entries of the feature matrix, so as to select discriminative inputs at both the node and feature-dimension levels:

$$X_M^{\gamma} = X \odot M_X^{\gamma} \tag{6}$$

where $\odot$ denotes the Hadamard product, and $M_X^{\gamma}$ denotes the mask matrix corresponding to the feature matrix X in the γ-th view. In this way, if the mask value $M_{X(i,j)}^{\gamma}$ corresponding to the j-th feature of the i-th brain region is small, the contribution of the feature value $X_{(i,j)}$ to the subsequent graph convolution process will be explicitly suppressed.

Second, at the structural connection level, this study does not assign independent learnable parameters to all edges. Instead, the edge mask is generated based on the joint decision of a weight matrix and node features. For a candidate edge (i,j) in the v-th view, its edge mask is defined as:

$$M_A^{\gamma} = \text{sigmoid}([feat_{v_i},\ feat_{v_j}] * W_A^{\gamma}) \tag{7}$$

where $M_A^{\gamma} \in R^{N \times N}$ denotes the edge mask matrix of the γ-th view, $feat_{v_i}$ and $feat_{v_j}$ denote the node features of node $v_i$ and node $v_j$, and $W_A^{\gamma} \in R^{N \times N}$ is the edge-weight transition mask matrix. The final edge-weight mask matrices $M_A^f$ and $M_A^s$ are jointly determined by the node features, thereby enabling joint modeling of features and structures. This feature-driven structural selection strategy means that the weights of different edges are dynamically generated by the same set of parameters according to the endpoint node features, rather than being learned independently for each edge. This not only reduces the number of parameters, but also maintains consistency between structural selection and node attributes.

Finally, the feature mask matrix and edge-weight mask matrices are applied to the multi-view graph G through Hadamard products:

$$\begin{cases} X_M^{\gamma} = X \odot M_X^{\gamma} \\ A_M^K = A^K \odot M_A^K \\ A_M^S = A^S \odot M_A^S \end{cases} \tag{8}$$

The node mask is used to learn the importance of different feature dimensions, while the edge mask is used to estimate the task-related importance of connections between ROIs. By assigning adaptive weights to node features and ROI-level candidate connections, enabling the model to focus on different brain connectivity patterns in different views. The key idea of this mechanism is that the multi-view masking mechanism does not simply assign global weights to different views; instead, it independently learns mask parameters within each view. This design allows the model to adaptively adjust its focus according to the structural characteristics of each view, thereby fully exploiting the complementarity among multiple views while avoiding the negative influence caused by the accumulation of redundant information across views.

**2.4 Cross-View Gated Fusion Module**

The structure of the cross-view gated fusion mechanism is illustrated in Figure 3.

**Figure 3.** Overview of the cross-view gated fusion module

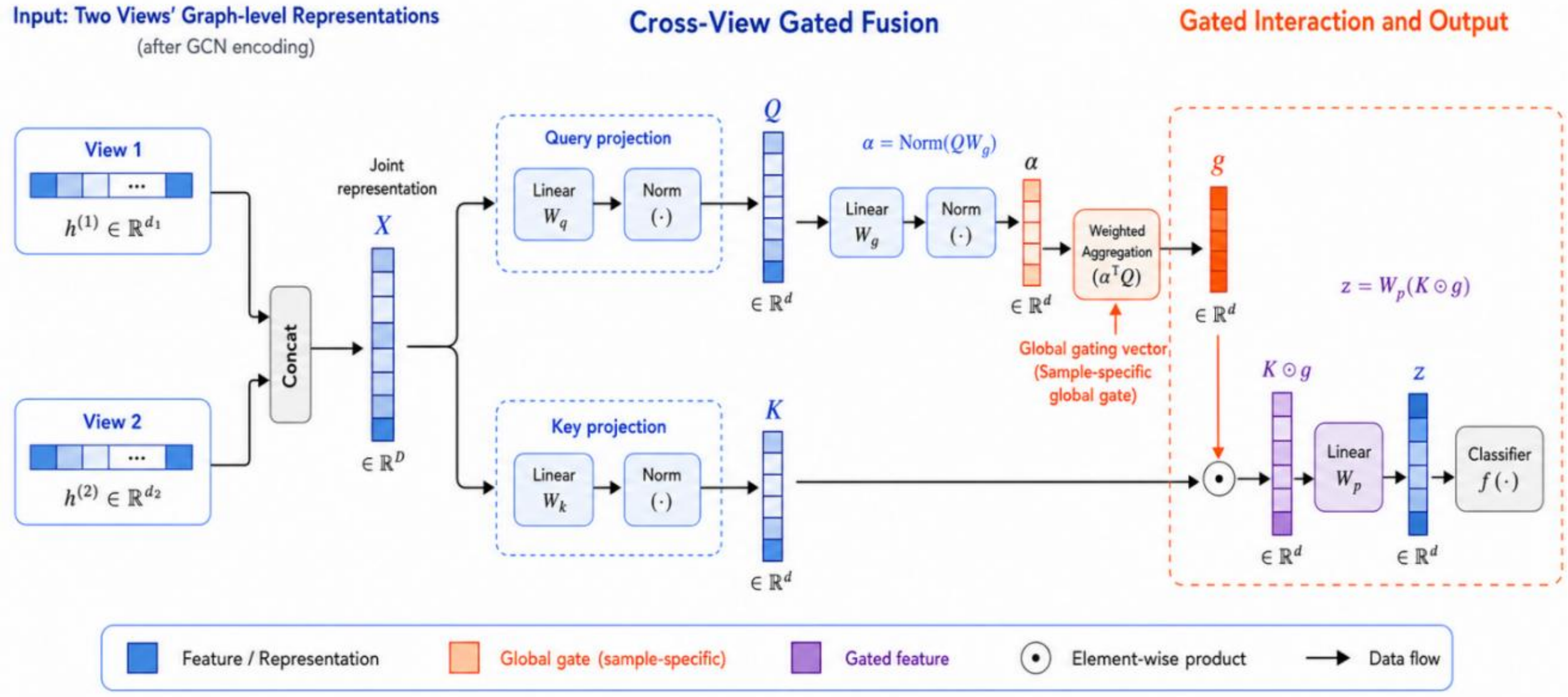

First, the graph-level features from all views are concatenated to form a joint representation U. Then, a query vector and a key vector are generated through linear mappings:：

$$Q= Norm(W_q \times U) \text{ , } K= Norm(W_k \times U) \tag{9}$$

where $W_q$ and $W_k$ are learnable parameters, and Norm( · ) denotes the normalization operation. Here, Q is used to characterize the response pattern of the current sample in the multi-view representation space, while K serves as the structural carrier for the subsequent gated interaction. On this basis, a gating weight vector $W_g$ is introduced to characterize the importance of different features. The response of the query vector along the gating direction is first calculated as follows:

$$\alpha=Norm(Q \times W_g), \quad g=\alpha^T Q \tag{10}$$

where α denotes the attention response induced by the query representation, and *g* denotes the global gating vector. Here, *g* does not correspond to any single view; instead, it serves as a compact representation of the overall discriminative focus of the current sample in the multi-view space. Compared with directly assigning fixed weights to individual views, the introduction of the gating vector prevents the fusion process from being limited to simple linear weighting and enables the model to capture, from a global perspective, the feature directions that should receive greater attention for the current sample.

Finally, the global gating vector is expanded to the same dimension as the key representation and interacts with the key representation through element-wise multiplication to obtain the final fused representation:

$$Z=W_p(K \odot g) \tag{11}$$

where ⊙ denotes the Hadamard product, $W_p$ is a linear projection matrix, and Z denotes the fused graph-level feature representation obtained after cross-view gated modulation. The proposed cross-view gated fusion mechanism differs from traditional Softmax-based attention weighting. Instead, it first extracts a global discriminative direction in the joint multi-view representation space and then uses this direction to modulate the key representation, thereby achieving adaptive reorganization of cross-view information.

After obtaining the fused graph-level representation Z, an MLP classifier is used to generate the final prediction logit:

$$p=\sigma(\mathrm{MLP}(Z)) \tag{12}$$

where p denotes the predicted probability. For a given sample, let $y \in \{0,1\}$ denote the ground-truth label and $\hat{p}$ denote the predicted probability after the sigmoid function. The classification loss is defined as:

$$\mathscr{L}_{cls}=-[y \log \hat{p} + (1-y) \log (1-\hat{p})] \tag{13}$$

**3、Experimental**

**3.1 Experimental Settings**

Experiments were conducted on the ADNI dataset. According to the clinical diagnostic labels, all samples with complete sMRI data were divided into three groups: Alzheimer's disease (AD), mild cognitive impairment (MCI), and cognitively normal (CN).

The ADNI database contains longitudinal follow-up data from cognitively normal (CN) individuals, patients with MCI, and patients with AD[25]. In this study, T1-weighted structural magnetic resonance imaging (T1-weighted sMRI) data were used. The final cohort in this study included 743 subjects, comprising 189 patients with AD, 304 CN subjects, and 250 patients with MCI, as shown in Table 1 and Table 2.

**Table 1. Demographic and imaging characteristics of the ADNI cohort.**

| Label | Age (years) | F/M | No. of Subjects | No. of sMRI Scans |
|---|---|---|---|---|
| AD | 75.13±7.62 | 79/110 | 294 | 971 |
| CN | 74.78±5.84 | 149/155 | 320 | 1656 |
| MCI | 74.39±7.34 | 85/165 | 294 | 1244 |

**Table 2. Training and testing distributions for different pairwise classification tasks.**

| Label | No. of Training/Testing Subjects | | | No. of Training/Testing sMRI Scans | | |
|---|---|---|---|---|---|---|
| | AD vs CN | AD vs MCI | CN vs MCI | AD vs CN | AD vs MCI | CN vs MCI |
| AD | 236/58 | 236/58 | \ | 785/186 | 796/175 | \ |
| CN | 256/64 | \ | 256/64 | 1331/325 | \ | 1331/325 |
| MCI | \ | 236/58 | 236/58 | \ | 985/259 | 979/265 |

For all medical images from all subjects, considering that the same subject may have multiple scans, subject-level data splitting was adopted to avoid information leakage. Specifically, all images from the same subject within the same diagnostic label were assigned exclusively to either the training set or the test set. Finally, 80% of the data were used for training and 20% for testing.

In terms of evaluation metrics, accuracy, precision, sensitivity, and F1-score were selected as performance evaluation criteria to comprehensively assess the classification ability of the model from multiple perspectives.

In terms of implementation, the proposed model was implemented based on PyTorch and PyTorch Geometric. The Adam optimizer was used, with a learning rate of 0.001, a training epoch number of 200, and a batch size of 32. The hidden layer dimensions for the three classification tasks were set to 64, 128, and 32, respectively. The random seed was set to 1234.

**3.2 Comparison with Advanced Methods**

To validate the effectiveness of the proposed method, we compared it with multiple baselines, including ablation variants and recent graph neural network-based approaches covering traditional GNNs, interpretable models, multi-view methods, and multimodal learning methods. To ensure fairness, all methods were evaluated under identical data splits, input features, evaluation metrics, training epochs, and optimizer settings, while model-specific hyperparameters followed the configurations reported in the original studies. For single-view methods, the feature similarity view constructed from sMRI data was used as input, whereas multi-view methods adopted the same two-view setting as the proposed framework. All methods were evaluated over five independent random runs, with mean performance and standard deviations reported for comparison. The compared methods are described as follows:

(1) BAGNN[26]. BAGNN, namely Brain-Aware Graph Neural Network, introduces a brain-aware readout layer to enhance the model's ability to capture key brain regions during graph-level representation learning, thereby improving the performance of early Alzheimer's disease detection.

(2) ADMGCN[27]. ADMGCN integrates a meta-learning strategy to improve the generalization ability of the model on small-sample medical datasets by sharing knowledge across different tasks, making it suitable for AD diagnosis scenarios.

(3) MVGNN[28]. MVGNN, is a typical multi-view graph neural network method. It enhances and aggregates multi-view structures to fuse information from different views, enabling effective modeling of multi-source structural relationships in brain networks.

(4) BrainGNN[29]. BrainGNN is an interpretable brain graph neural network model. By introducing ROI selection and subgraph learning mechanisms, it can automatically identify key brain regions and provides good interpretability in brain disease analysis tasks.

(5) FIL[30]. The FIL method combines feature inductive learning with a multilayer graph neural network structure to jointly model multimodal data, showing strong feature representation capability in Alzheimer's disease diagnosis tasks.

The experimental results are shown in Table 3. For the AD–CN task, the proposed method achieved the most prominent advantage. Compared with the best-performing baseline method in this task, RSEA-MVGNN, the proposed method improved Accuracy from 82.06% to 89.90%, F1-score from 75.02% to 86.24%, and Sensitivity from 74.73% to 86.77%. These results indicate that, for the AD–CN task where structural differences are relatively more pronounced, multi-view selection and cross-view gating can effectively exploit discriminative patterns in brain networks.

In the more challenging AD–MCI and CN–MCI tasks, the proposed method also showed stable advantages. For AD–MCI classification, the proposed method achieved an Accuracy of 72.12% and an F1-score of 68.41%, outperforming the best baseline by 1.38 and 3.22 percentage points, respectively. In particular, its Sensitivity reached 74.97%, representing an improvement of 9.26 percentage points over the best baseline, indicating that the model is more sensitive to early pathological samples.

For CN–MCI classification, the proposed method achieved an Accuracy of 69.70% and an F1-score of 68.16%, improving upon the best baseline by 2.51 and 3.39 percentage points, respectively. Although some baseline methods showed local advantages in individual metrics, such as higher Specificity or Sensitivity, the proposed method achieved a better overall balance across most tasks. This suggests that the proposed model does not simply favor one class but possesses more robust overall discriminative capability in scenarios with complex class boundaries.

**Table 3.** Performance comparison with representative graph neural network-based methods

| Task | Model | Accuracy | F1 Score | Precision | Sensitivity |
|---|---|---|---|---|---|
| AD vs CN | BAGNN | 0.7724±0.0079 | 0.6780±0.0159 | 0.6988±0.0112 | 0.6591±0.0286 |
| | ADMGCN | 0.7894±0.0122 | 0.7210±0.0088 | 0.6999±0.0372 | 0.7473±0.0391 |
| | MVGNN | 0.8206±0.0103 | 0.7502±0.0173 | 0.7603±0.0167 | 0.7414±0.0331 |
| | FIL | 0.8173±0.0120 | 0.7476±0.0206 | 0.7794±0.0268 | 0.7202±0.0436 |
| | Brain GNN | 0.8165±0.0163 | 0.7474±0.0247 | 0.7482±0.0213 | 0.7473±0.0355 |
| | MVMGNN | **0.8990±0.0112** | **0.8624±0.0124** | **0.8585±0.0368** | **0.8677±0.0230** |
| AD vs MCI | BAGNN | 0.6996±0.0149 | 0.6235±0.0152 | 0.6331±0.0316 | 0.6183±0.0459 |
| | ADMGCN | 0.7009±0.0153 | 0.6374±0.0323 | 0.6236±0.0198 | 0.6571±0.0739 |
| | MVGNN | 0.7074±0.0099 | 0.6233±0.0154 | 0.6505±0.0277 | 0.6023±0.0467 |
| | FIL | 0.6322±0.0338 | 0.6519±0.0857 | 0.6606±0.0469 | 0.6527±0.0857 |
| | Brain GNN | 0.7005±0.0089 | 0.6250±0.0111 | 0.6317±0.0161 | 0.6194±0.0247 |
| | MVMGNN | **0.7212±0.0093** | **0.6841±0.0071** | **0.6315±0.0253** | **0.7497±0.0458** |
| CN vs MCI | BAGNN | 0.6554±0.0117 | 0.6167±0.0170 | 0.6173±0.0196 | 0.6185±0.0417 |
| | ADMGCN | 0.6414±0.0176 | 0.6261±0.0156 | 0.5916±0.0296 | 0.6702±0.0525 |
| | MVGNN | 0.6719±0.0105 | 0.6260±0.0074 | 0.6420±0.0177 | 0.6113±0.0114 |
| | FIL | 0.6501±0.0198 | 0.6477±0.0151 | 0.5769±0.0216 | **0.7400±0.0354** |
| | Brain GNN | 0.6654±0.0045 | 0.6333±0.0152 | 0.6251±0.0195 | 0.6453±0.0473 |
| | MVMGNN | **0.6970±0.0138** | **0.6816±0.0069** | **0.6471±0.0275** | 0.7223±0.0347 |

### 3.3 Ablation Results

To systematically analyze the influence of each key module on model performance, this study conducted ablation experiments to progressively verify the effects of the multi-view structure, the masking mechanism, and the cross-view gated fusion mechanism. Six model variants were constructed as follows:

(1) A comparison between single-view and dual-view models was conducted to verify the effectiveness of the multi-view structure.

(2) A comparison between the basic dual-view model and the model with the masking mechanism was conducted to verify the role of the joint node–edge masking mechanism.

(3) A comparison between models with and without gated fusion was conducted to verify the effect of the cross-view gated fusion mechanism.

(4) The complete model was compared with all variants to analyze the synergistic effects of multiple modules.

As shown in Table 4, the ablation results demonstrate the contributions of multi-view modeling, the joint node–edge masking mechanism, and the cross-view gated fusion mechanism.

For the AD–CN task, the basic dual-view model achieved an Accuracy of 82.78% and an F1-score of 76.57%, indicating that the spatial view and the feature similarity view provide complementary information. After introducing the masking mechanism, the Accuracy and F1-score increased to 88.92% and 85.26%, respectively, while Sensitivity improved from 77.31% to 88.06%. This suggests that intra-view selection can effectively highlight key brain regions and structural connections closely related to AD. After further incorporating cross-view gated fusion, the complete model improved the Accuracy and F1-score to 89.90% and 86.24%, respectively.

In the AD–MCI task, the class boundary is more ambiguous, and the benefits brought by different modules show more distinct characteristics. The basic dual-view model achieved an Accuracy of 72.90% and an F1-score of 65.80%. When only cross-view gated fusion was introduced, Sensitivity increased from 64.80% to 67.54%, while Accuracy slightly decreased to 72.21%. When only the masking mechanism was introduced, the F1-score improved from 65.80% to 67.75%, and Sensitivity further increased to 76.00%. This indicates that the masking mechanism tends to enhance the detection of pathological samples. Based on this, the complete model achieved the highest F1-score of 68.41% while maintaining a relatively high Sensitivity of 74.97%.

The CN–MCI task is also highly challenging. The basic dual-view model achieved an Accuracy of 66.10% and an F1-score of 63.03%. When only cross-view gated fusion was used, Sensitivity increased to 77.21%. When only the masking mechanism was used, Accuracy improved to 67.93%. The complete model finally achieved the best performance on this task, with an Accuracy of 69.70% and an F1-score of 68.16%. These results indicate that the masking mechanism and the gating mechanism are not simply additive; instead, they act on intra-view selection and inter-view fusion, respectively, showing clear complementarity.

Overall, the ablation experiments demonstrate that multi-view modeling provides complementary structural information, the masking mechanism mainly enhances the selection of key features and connections, and cross-view gated fusion further alleviates the imbalance in contributions from different views. The synergistic effect of these three components enables the complete model to achieve more stable and balanced discriminative performance across all three classification tasks.

**Table 4.** Ablation study of the proposed MVMGNN

| Task | KNN View | Similarity View | Node-Edge Mask | Gate Fusion | Accuracy | F1 Score | Precision | Sensitivity |
|---|---|---|---|---|---|---|---|---|
| AD vs CN | √ | | | | 0.8196±0.0026 | 0.7586±0.0072 | 0.7394±0.0086 | 0.7796±0.0231 |
| | | √ | | | 0.8356±0.0035 | 0.7807±0.0052 | 0.7593±0.0155 | 0.8043±0.0255 |
| | √ | √ | | | 0.8278±0.0055 | 0.7657±0.0061 | 0.7587±0.0118 | 0.7731±0.0079 |
| | √ | √ | | √ | 0.8301±0.0080 | 0.7712±0.0117 | 0.7567±0.0153 | 0.7871±0.0247 |
| | √ | √ | √ | | 0.8892±0.0042 | 0.8526±0.0048 | 0.8278±0.0232 | 0.8806±0.0283 |
| | √ | √ | √ | √ | **0.8990±0.0112** | **0.8624±0.0124** | **0.8585±0.0368** | **0.8677±0.0230** |
| AD vs MCI | √ | | | | 0.7069±0.0079 | 0.6174±0.0154 | 0.652±0.0139 | 0.5874±0.0305 |
| | | √ | | | 0.6953±0.0102 | 0.6156±0.022 | 0.6268±0.0147 | 0.6068±0.0444 |
| | √ | √ | | | 0.7290±0.0094 | 0.6580±0.0145 | 0.6714±0.0238 | 0.6480±0.0432 |
| | √ | √ | | √ | 0.7221±0.0168 | 0.6619±0.0069 | 0.6552±0.0420 | 0.6754±0.0509 |
| | √ | √ | √ | | 0.7083±0.0136 | 0.6775±0.0061 | 0.6128±0.0211 | 0.7600±0.0337 |
| | √ | √ | √ | √ | **0.7212±0.0093** | **0.6841±0.0071** | **0.6315±0.0253** | **0.7497±0.0458** |
| CN vs MCI | √ | | | | 0.6407±0.0066 | 0.6073±0.0185 | 0.5976±0.0168 | 0.6211±0.0499 |
| | | √ | | | 0.6474±0.0063 | 0.6201±0.0221 | 0.6019±0.0155 | 0.6445±0.0648 |
| | √ | √ | | | 0.6610±0.0051 | 0.6303±0.0077 | 0.6181±0.0100 | 0.6438±0.0232 |
| | √ | √ | | √ | 0.6332±0.0142 | 0.6537±0.0100 | 0.5688±0.0173 | 0.7721±0.0484 |
| | √ | √ | √ | | 0.6793±0.0130 | 0.6459±0.0079 | 0.6424±0.0251 | 0.6513±0.0263 |
| | √ | √ | √ | √ | **0.6970±0.0138** | **0.6816±0.0069** | **0.6471±0.0275** | **0.7223±0.0347** |

### 3.4 External Validation on Independent Datasets

To further evaluate the applicability and robustness of the proposed MVMGNN on independent datasets, external dataset validation experiments were conducted on two public sMRI datasets, OASIS and AIBL. These two datasets were not involved in the training or testing process on the ADNI dataset, and were therefore used to assess the reproducibility of the proposed modeling strategy across different cohorts and imaging data sources.

The OASIS and AIBL databases contain structural MRI data from cognitively normal (CN) individuals, patients with mild cognitive impairment (MCI), and patients with Alzheimer’s disease (AD). In this study, T1-weighted structural magnetic resonance imaging (T1-weighted sMRI) data were used for external dataset evaluation. The cohort information and diagnostic group distributions of the OASIS and AIBL datasets are summarized in Table 5.

**Table 5.** Cohort characteristics and diagnostic distribution of the AIBL and OASIS datasets.

| Dataset | Label | Age (years) | F/M | No. of Subjects | No. of sMRI Scans |
|---|---|---|---|---|---|
| AIBL | AD | 75.03±7.29 | 41/35 | 76 | 105 |
| | CN | 72.02±6.28 | 110/79 | 189 | 230 |
| | MCI | 74.97±6.86 | 42/48 | 90 | 121 |
| OASIS | AD | 76.57±7.61 | – | 92 | 116 |
| | CN | 70.86±7.65 | – | 99 | 116 |
| | MCI | 73.55±6.92 | – | 69 | 77 |

The classification results on the OASIS and AIBL datasets are reported in Table 6. Overall, the proposed method achieved relatively stable classification performance on both external datasets, suggesting that the multi-view graph construction, joint node–edge masking mechanism, and cross-view gated fusion module can be effectively applied to independent sMRI cohorts. Although differences in subject composition, scanning protocols, sample size, and diagnostic distribution may lead to performance variations across datasets, the external dataset validation results indicate that MVMGNN still maintains favorable discriminative ability and cross-dataset applicability beyond the ADNI dataset.

**Table 6.** External validation results on the OASIS and AIBL datasets.

| Dataset | Task | Accuracy | F1 Score | Precision | Sensitivity |
|---|---|---|---|---|---|
| AIBL | AD vs CN | 0.8746 ± 0.0170 | 0.7967 ± 0.0356 | 0.8517 ± 0.0512 | 0.7546 ± 0.0760 |
| | AD vs MCI | 0.7116 ± 0.0127 | 0.6923 ± 0.0181 | 0.7672 ± 0.0553 | 0.6364 ± 0.0556 |
| | CN vs MCI | 0.7342 ± 0.0315 | 0.6503 ± 0.0286 | 0.6627 ± 0.0624 | 0.6429 ± 0.0438 |
| OASIS | AD vs CN | 0.8038 ± 0.0179 | 0.7881 ± 0.0234 | 0.7893 ± 0.0392 | 0.7901 ± 0.0513 |
| | AD vs MCI | 0.7143 ± 0.0036 | 0.7416 ± 0.0108 | 0.7500 ± 0.0185 | 0.7339 ± 0.0201 |
| | CN vs MCI | 0.6738 ± 0.0502 | 0.5669 ± 0.0992 | 0.5980 ± 0.0907 | 0.5580 ± 0.1463 |

### 3.5 Interpretability Analysis

To further evaluate the interpretability of the proposed model, we visualized the learned node importance and edge importance for the three classification tasks. In the proposed model, the node importance score of each ROI was derived from the learned node feature masks by averaging the mask values across radiomics feature dimensions. The edge importance score was derived from the learned edge masks, which reflect the task-related contribution of candidate structural connections between ROI nodes. For visualization, the learned importance scores were normalized to the range of 0 to 1.

As shown in Figure. 4(a), the node importance maps reveal that the discriminative patterns identified by the model are not restricted to a single focal region, but are distributed across multiple cortical and subcortical areas. Moreover, the spatial distributions of important nodes vary across the AD/CN, AD/MCI, and CN/MCI tasks, suggesting that the model captures task-specific structural abnormality patterns associated with different disease stages.

Figure. 4(b) further presents the edge importance matrices learned by the model. The heatmaps exhibit non-uniform connection patterns, indicating that the model assigns different levels of importance to different inter-regional structural relationships rather than treating all edges equally. In addition, the edge importance distributions differ across the three tasks, which implies that the discriminative structural interactions among brain regions are also task-dependent.

To provide a more intuitive anatomical interpretation of the learned structural interaction patterns, Figure. 4(c) further presents circular brain connectivity diagrams of important ROI-level connections. In each diagram, the outer ring represents ROI-level node importance, where warmer colors indicate higher node importance. The inner green chords represent important structural connections identified by the model, and thicker chords indicate higher edge importance.

Overall, the visualization results show that the proposed model can simultaneously identify informative brain regions and important structural connections. These findings suggest that the model learns biologically meaningful and task-relevant structural representations, thereby improving not only classification performance but also the interpretability of the learned decision patterns.

**Figure 4.** Visualization of node and edge importance learned by MVMGNN: (a) Node importance maps for the three classification tasks. Highlighted regions indicate ROIs assigned higher importance by the learned node masks; (b) Edge importance matrices for the three classification tasks. Higher values indicate stronger task-related contributions of inter-regional structural connections; (c) Circular visualization of ROI-level node and edge importance. The outer ring represents ROI-level node importance, and the inner green chords represent important structural connections, with thicker chords indicating higher edge importance.

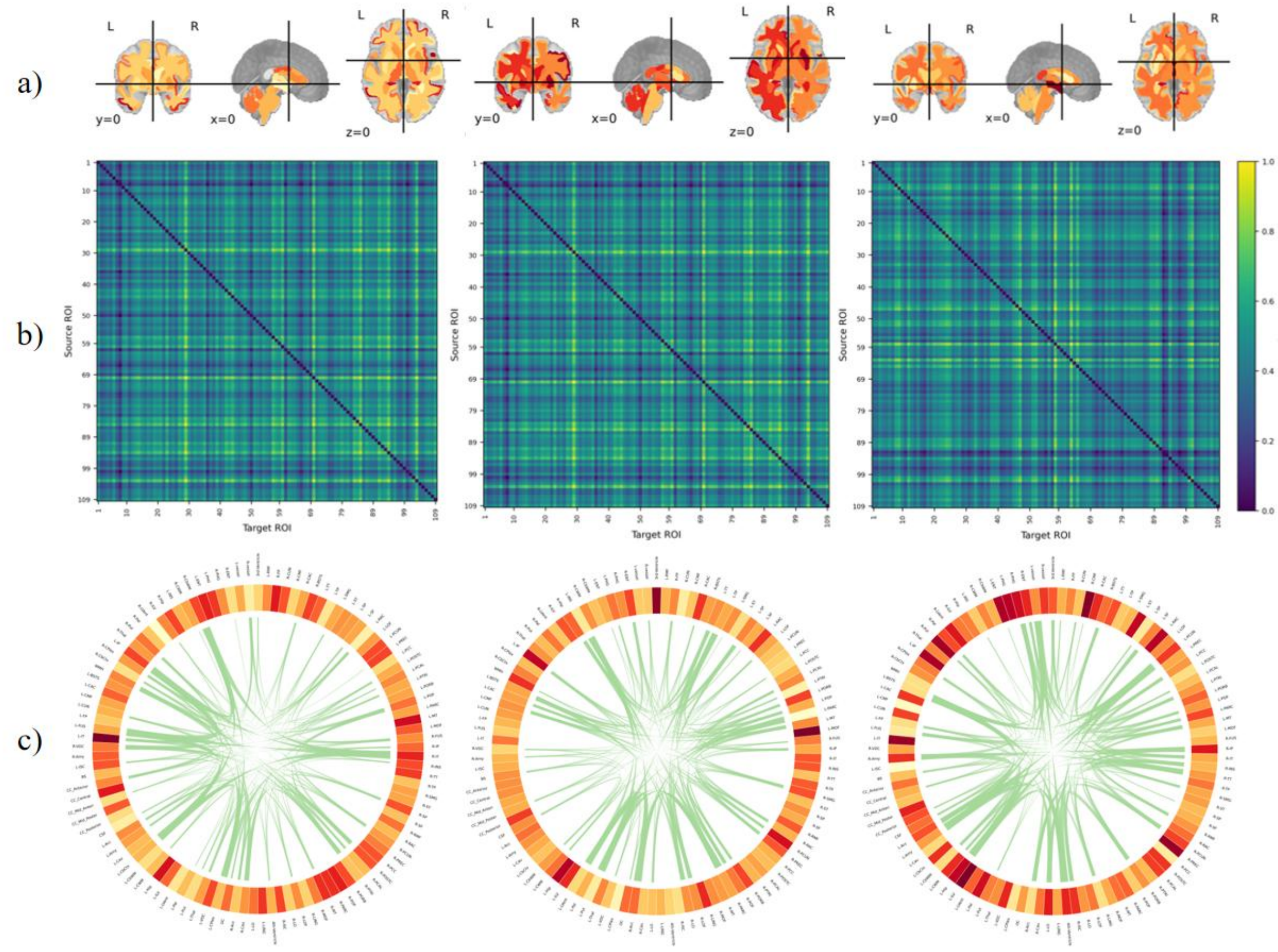

**4. Discussion**

**4.1 Main Contributions**

This study proposes MVMGNN for sMRI-based AD-related classification. The main contribution of the proposed model lies in its ability to jointly characterize ROI-level radiomics features and inter-regional structural relationships from complementary graph views. Specifically, the spatial proximity view describes local anatomical neighborhood relationships among brain regions, whereas the feature similarity view captures morphological associations in the radiomics feature space. By integrating these two views, the model can provide a more comprehensive representation of structural brain organization than conventional single-view graph construction strategies.

Another important contribution of this study is the introduction of a joint node–edge masking mechanism. Multi-view graph construction may introduce redundant node features and noisy candidate connections, which can interfere with graph convolutional propagation if directly used for message passing. To address this issue, the proposed masking mechanism performs intra-view information selection from both the node feature dimension and the structural connection level. The node mask learns the importance of different radiomics feature dimensions, while the edge mask estimates the task-related importance of candidate structural connections between ROIs. The ablation results demonstrate that the masking mechanism improves model performance across the three classification tasks, indicating that task-oriented intra-view selection is beneficial for learning discriminative structural brain network representations.

In addition, this study introduces a patient-level cross-view gated fusion mechanism for adaptive multi-view integration. The gated fusion module modulates the joint multi-view representation according to subject-specific structural patterns. The ablation results show that the complete model achieves further improvements after combining joint node–edge masking with cross-view gated fusion, suggesting that intra-view selection and inter-view adaptive fusion play complementary roles. Overall, the experimental results on the ADNI dataset demonstrate that MVMGNN achieves favorable performance in the AD-CN, AD-MCI, and CN-MCI classification tasks.

### 4.2 Performance and Interpretability Analysis

The effectiveness of the proposed model can be further analyzed from two aspects: classification performance across different diagnostic tasks and the interpretability of the learned brain network patterns. From the perspective of different classification tasks, the performance gains of the proposed method are closely related to task difficulty. In the AD-CN task, the structural differences between the two classes are relatively more pronounced, allowing the model to fully exploit complementary multi-view information and achieve more significant performance improvements. In contrast, the AD-MCI and CN-MCI tasks are more challenging because the disease stages are closer and the class boundaries are more ambiguous. Nevertheless, the proposed method still shows certain advantages in metrics such as F1-score and sensitivity, suggesting that the model has good recognition capability for early-stage or boundary-ambiguous samples.

In terms of interpretability, the node masking analysis shows that brain regions assigned higher weights by the model include the hippocampus and temporal lobe-related regions. These regions are generally consistent with structural abnormality regions reported in previous neuroimaging studies of AD, indicating that the discriminative patterns learned by the model have certain neurobiological plausibility. Meanwhile, the high-weight brain regions are not concentrated in a single local area but are distributed across multiple cortical and subcortical structures, suggesting that the discriminative information captured by the model may involve a relatively widespread distribution of brain regions rather than relying solely on a single brain region for classification. It should be noted that the masking weights reflect the model’s reliance on different brain region features and connections during classification, and they cannot be directly interpreted as causal biomarkers in the disease mechanism.

### 4.3 Limitations and Future Work

Although the proposed method achieves promising experimental results, several limitations remain. First, this study only models structural MRI data and does not incorporate other modalities such as PET, fMRI, genetic information, or clinical scales. Therefore, its ability to characterize the complex pathological mechanisms and individual heterogeneity of AD remains limited. Future work may further integrate multimodal neuroimaging and clinical features to construct a more comprehensive multi-source diagnostic framework. Second, the multi-view construction in this study relies on predefined spatial proximity and feature similarity relationships. Although these two types of views provide complementary information, their optimality still requires further validation. Future studies may explore learnable adjacency matrices, dynamic graph structure learning, or adaptive graph construction strategies to improve the flexibility and generalization ability of the model.

In addition, the experiments in this study were mainly conducted on the ADNI dataset. Although repeated experiments were performed to reduce the uncertainty caused by random data partitioning, the generalization ability of the model on external independent datasets still needs further validation. Future work may test the proposed method on more multicenter and independent cohorts, and incorporate longitudinal follow-up data to evaluate its ability to predict disease progression and the risk of MCI conversion. Overall, this study provides a multi-view graph learning framework that balances discriminative performance and interpretability for sMRI-based structural brain network analysis in the computer-aided diagnosis of Alzheimer's disease. It also provides a foundation for future research on multimodal fusion, adaptive graph construction, and clinical translation.

## 5. Conclusion

This study addresses the limitations of insufficient single-view representation and redundant multi-view information in sMRI-based structural brain network modeling by proposing a graph neural network model incorporating multi-view masking and cross-view gated fusion mechanisms. The proposed method constructs structural brain networks from two perspectives, namely spatial proximity and feature similarity, to introduce complementary inter-regional relationship information. Meanwhile, a joint node–edge masking mechanism is designed within each view to adaptively select brain-region radiomics features and candidate structural connections, thereby suppressing the influence of redundant features and noisy connections during graph convolutional propagation. On this basis, a sample-level cross-view gated fusion mechanism is further introduced to achieve adaptive integration of graph-level representations from different views.

Experimental results on the ADNI dataset demonstrate that the proposed method achieves superior overall performance compared with multiple competing methods in the AD-CN, AD-MCI, and CN-MCI classification tasks, validating the effectiveness of multi-view modeling, joint node–edge masking, and cross-view gated fusion. Interpretability analysis further indicates that the model can identify key brain regions associated with AD, providing useful insights into the discriminative patterns embedded in sMRI-based brain networks.

Future work will further incorporate multimodal information, such as PET, fMRI, and clinical assessment scales, to enhance the model's ability to characterize AD heterogeneity and disease progression. In addition, more adaptive graph construction strategies and more efficient feature selection mechanisms will be explored to improve the generalization ability and practical value of the model in independent datasets and real clinical scenarios.

Funding:

Ni Yao received support from the Natural Science Foundation of Henan Province (Grant Number: 262300421818) and the Provincial Key Research Project of higher education institutions in Henan (Grant Number: 26A520044), Danyang Sun received support from the National Natural Science Foundation of China(Grant Numbers: 62506343) and the International Science and Technology Cooperation Project of Henan Province (Grant Number: 262102521008), Chuang Han received support from the Zhongyuan Science and the Technology Innovation Outstanding Young Talents Program, Yanting Li received support from the Science and Technology Innovation Talent Project of Henan Province University (Grant Number: 25HASTIT028), Fubao Zhu received support from the National Natural Science Foundation of China (Grant Numbers: 62476255, and 82370513).